\documentclass[10pt,twocolumn,letterpaper]{article}

\usepackage{iccv}
\usepackage{times}
\usepackage{epsfig}
\usepackage{graphicx}
\usepackage{amsmath}
\usepackage{amssymb}
\usepackage{paralist}
\usepackage[sort,numbers]{natbib}
\usepackage{booktabs}
\usepackage{bbold}
\usepackage{subcaption}
\usepackage{color, xcolor}
\usepackage{ragged2e} 
\usepackage{makecell, multirow, tabularx}
\usepackage{threeparttable}

\usepackage{authblk}

\usepackage[accsupp]{axessibility}  

\usepackage[pagebackref,breaklinks,colorlinks]{hyperref}
\usepackage[capitalize]{cleveref}
\crefname{section}{Sec.}{Secs.}
\Crefname{section}{Section}{Sections}
\Crefname{table}{Table}{Tables}
\crefname{table}{Tab.}{Tabs.}

\iccvfinalcopy 

\def\iccvPaperID{4976} 

\ificcvfinal\fi

\begin{document}

\title{Towards Deeply Unified Depth-aware Panoptic Segmentation \\with Bi-directional Guidance Learning}


\author[1]{Junwen He}
\author[1]{Yifan Wang}
\author[1]{Lijun Wang\thanks{Corresponding author}}
\author[1]{Huchuan Lu}
\author[2]{Jun-Yan He}
\author[2]{\\ Jin-Peng Lan}
\author[2]{Bin Luo}
\author[2]{Yifeng Geng}
\author[2]{Xuansong Xie}

\affil[1]{Dalian University of Technology}
\affil[2]{DAMO Academy, Alibaba Group}



\maketitle
\ificcvfinal\thispagestyle{empty}\fi

\begin{abstract}
Depth-aware panoptic segmentation is an emerging topic in computer vision which combines semantic and geometric understanding for more robust scene interpretation.
Recent works pursue unified frameworks to tackle this challenge but mostly still treat it as two individual learning tasks, which limits their potential for exploring cross-domain information.
We propose a deeply unified framework for depth-aware panoptic segmentation, which performs joint segmentation and depth estimation both in a per-segment manner with identical object queries. 
To narrow the gap between the two tasks, we further design a geometric query enhancement method, which is able to integrate scene geometry into object queries using latent representations. 
In addition, we propose a bi-directional guidance learning approach to facilitate cross-task feature learning by taking advantage of their mutual relations.
Our method sets the new state of the art for depth-aware panoptic segmentation on both Cityscapes-DVPS and SemKITTI-DVPS datasets.
Moreover, our guidance learning approach is shown to deliver performance improvement even under incomplete supervision labels.
Code and models are available at \url{https://github.com/jwh97nn/DeepDPS}.
\end{abstract}

\section{Introduction}

Scene understanding plays a crucial role in autonomous driving perception systems, but relying solely on 2D representations falls short for advanced systems.
To address this limitation, Depth-aware Panoptic Segmentation (DPS) \cite{qiao2021vip} has been proposed as a novel approach for geometric scene understanding, which enables the creation of 3D instance-level semantic labels from a single image by means of inverse projection.
More precisely, the simplified problem can be decomposed into two sub-tasks: panoptic segmentation and monocular depth estimation.

\begin{figure}[t]
\centering
\includegraphics[width=\columnwidth]{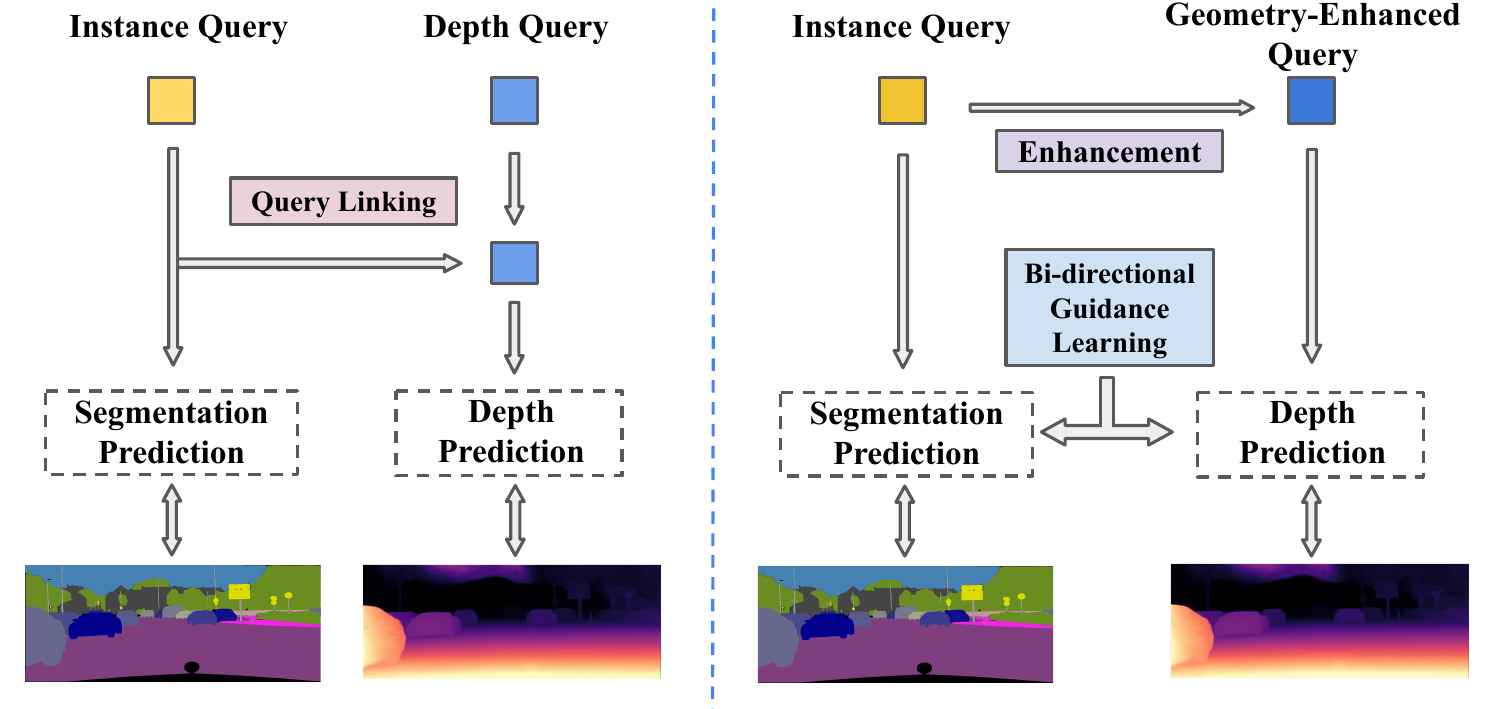}
\caption{Pipeline comparison of prior work \cite{yuan2021polyphonicformer} (left) and ours (right). We integrate unified queries with geometric enhancement and mutual learning from cross-modality supervision, towards a deeper unified manner.}
\label{1}
\end{figure}

Early methods \cite{schon2021mgnet, qiao2021vip} tackle this task by simply attaching a dense depth prediction head on top of the off-the-shelf panoptic segmentation model \cite{cheng2020panopticdeeplab}.
However, these methods are intuitively sub-optimal, because the separate task-oriented head design treats these two sub-tasks independently and ignores their mutual relation.
Recent methods \cite{gao2022panopticdepth, yuan2021polyphonicformer} propose unified architectures that output both predictions in the same instance-wise manner, and utilize corresponding task-specific kernels (or queries) to jointly produce masks and depth maps for individual instances, which leverages the mutual benefits between semantic and depth information (shown in \cref{1} left).

Despite recent efforts to unify the two sub-tasks, their learning processes remain largely separate. 
Specifically, they employ task-specific loss functions to guide individual predictions, which overlook the potential benefits of cross-domain knowledge learning.
While some attempts \cite{guizilini2020semantically, jung2021fine} have been done to learn depth representations from semantic segmentation implicitly, the reciprocal relationship between the two tasks remains largely unexplored.

In this study, we introduce a new deeply unified framework for depth-aware panoptic segmentation, which leverages cross-modality knowledge not only at the architectural level but also during the learning phase.
Rather than using separate queries for each task, we employ unified queries followed by geometry enhancement with latent representations.
Furthermore, we design a bi-directional guidance learning approach to optimize multi-task feature learning, which can better leverage their interdependence by using the supervision of one to guide the other.

We propose a deeply unified encoder-decoder architecture, which performs joint panoptic segmentation and depth estimation in a per-segment manner with identical queries.
We first generate instance-specific masks using unified per-segment queries, and enhance the queries with intermediate depth features as well as learned latent representations to integrate scene geometry, via the proposed Geometric Query Enhancement (\cref{depthquery}).
Subsequently, we predict depth maps from each enhanced query via a dot product with the depth embedding, and apply corresponding mask predictions to produce instance-wise depth predictions.
To account for low-confidence filtering, which causes imperfect masks or blank segments, we introduce an extra backup query to cover up these regions.

Moreover, we present a novel approach to leverage cross-modality knowledge by refining intermediate feature representations through Bi-directional Guidance Learning.
Our approach is based on the intuition that pixels crossing the semantic boundary are more likely to have a significant difference in depth and vice versa.
To this end, we propose Semantic-to-Depth guidance to optimize relative depth feature distances using contrastive learning, and Depth-to-Semantic guidance to synchronize semantic feature continuity with depth annotations.
The combination of both guidance mechanisms enables us to exploit their deeply-coupled relations and promote a more mutually-beneficial learning process.

Our method makes the following contributions:
\begin{compactenum}
    \item We propose a new deeply unified architecture for depth-aware panoptic segmentation, which tackles both sub-tasks in a per-segment manner, by integrating scene geometry into unified queries with geometry enhancement.
    \item We propose a new training method that refines both intermediate features simultaneously through bi-directional guidance learning, leveraging their mutual relations and boosting performance under incomplete supervision.
    \item Extensive experiments on Cityscapes-DVPS \cite{qiao2021vip} and SemKITTI-DVPS \cite{qiao2021vip} demonstrate the effectiveness of our proposed method, leading to state-of-the-art performance on depth-aware panoptic segmentation and individual sub-tasks.
\end{compactenum}

\section{Related Works}

\subsection{Panoptic Segmentation}

Early methods for panoptic segmentation \cite{kirillov2019panoptic} usually employ a two-stage (or proposal-based) approach \cite{xiong2019upsnet, zhao2017pyramid, li2019attention, lazarow2020learning, lin2017feature, he2021mgseg}, which generates instance-level segmentation based on region proposals, followed by post-processing fusion \cite{kirillov2019panoptic, kirillov2019panopticfpn} to obtain the final panoptic segmentation results.
Instead, bottom-up (or box-free) approaches attempt to group pixels to generate instance masks on top of semantic segmentation results. 
For example, DeeperLab \cite{yang2019deeperlab} proposes to reformulate panoptic segmentation task into keypoint and multi-range offset heatmap predictions, with a fusion process followed by \cite{papandreou2018personlab}. 
Panoptic-DeepLab \cite{cheng2020panopticdeeplab} and its variants \cite{wang2020axialdeeplab, chen2020scaling} predict class-agnostic instance centers, together with pixel-level offsets to the corresponding center. 
More recently, unified architectures have been proposed and adopted by many studies, which surpass previous single-stage methods while avoiding complex post-processing.
Panoptic FCN \cite{li2021panopticfcn} encodes each instance or stuff into a specific kernel, achieving instance-aware and semantically consistent representations. 
K-Net \cite{zhang2021knet} proposes a kernel update strategy to iteratively refine the kernels and mask predictions. 
MaskFormer \cite{cheng2021maskformer} and Mask2Former \cite{cheng2022mask2former} reformulate the segmentation task into mask classification, and use the DETR-like \cite{carion2020detr} architecture to predict a set of binary masks from learned object queries. 
When extending the unified framework to the video level, \cite{kim2022tubeformer} and \cite{hwang2021ifc} incorporated a memory token to enable communication across multi-frame features.
Inspired by these works, we extend the unified architecture to perform instance-level depth estimation together with panoptic segmentation, by incorporating both features through the learned latent representation and leveraging their mutual relations.

\begin{figure*}[htb]
    \centering
    \includegraphics[width=\textwidth]{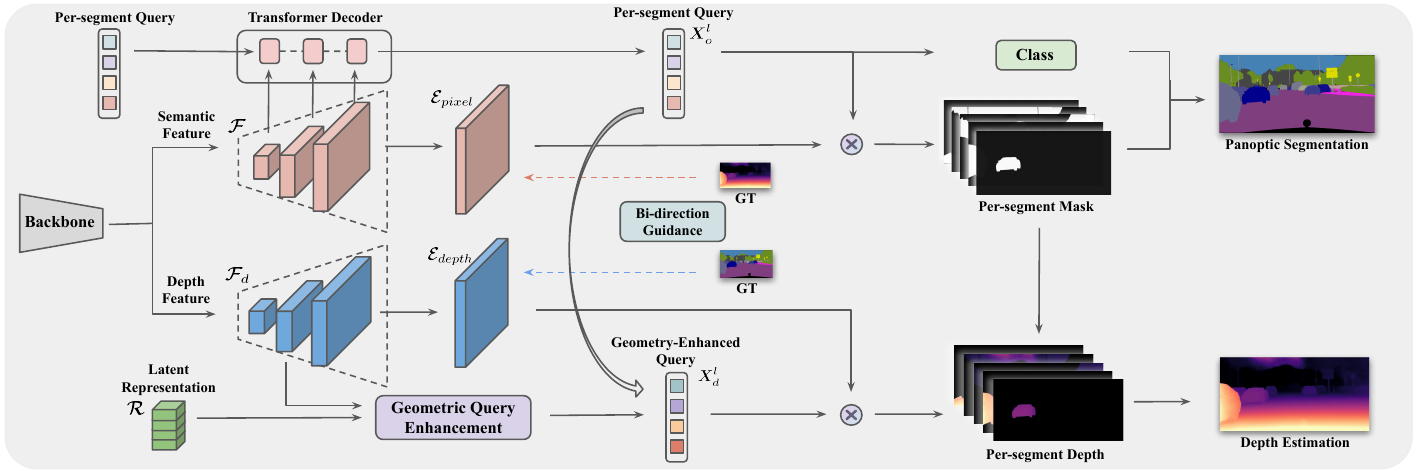}
    \caption{
    {\textbf{Architecture Overview.}} We learn unified per-segment queries $X_o^l$ and obtain geometry enhanced queries $X_d^l$, by incorporating multi-scaled depth features $\mathcal{F}_d$ and learned latent representations $\mathcal{R}$ through \textbf{Geometry Query Enhancement}.
    We introduce a \textbf{Bi-directional Guidance Learning} to refine both features with cross modality supervisions, which includes Semantic-to-Depth ({\textcolor[RGB]{109,158,235}{blue dashed arrow}}) and Depth-to-Semantic (\textcolor[RGB]{221,127,107}{orange dashed arrow}) guidance.
    $\otimes$ denotes dot product.}
    \label{network}
\end{figure*}

\subsection{Monocular Depth Estimation}

The estimation of depth from a single image is a challenging problem in 3D computer vision.
Eigen \textit{et al.}\cite{eigen2014depth} proposes the first learning-based method, which uses multi-scale CNNs to predict depth maps directly.
Follow-up methods exploit more powerful network architectures \cite{laina16, li17twostreamed, bhat21adabins, ranftl21dpt} or reformulate the task as a classification problem and predict a fixed \cite{fu18dorn} or adaptive \cite{bhat21adabins, li2022binsformer} range for each pixel.
Other methods perform multi-task learning (\textit{e.g.} surface normal \cite{li15normal, yin18geonet} and semantic labels \cite{eigen15common, hu19revisiting}) to enhance depth predictions. 
Among them, SDC-Depth \cite{wang2020sdc} decomposes the global depth prediction task into a series of category-specific ones, with the help of off-the-shelf segmentation module.
Our approach to depth estimation also involves predicting instance-wise depth maps.
In contrast, this is achieved through a unified model that generates both instance masks and depth maps, and is further boosted from mutual learning between modalities.

\subsection{Depth-aware Panoptic Segmentation}

The task of combining panoptic segmentation and monocular depth estimation was initially proposed in \cite{qiao2021vip}, but numerous studies have examined the relationship between these two tasks prior.
Earlier approaches have viewed this as a multi-task learning problem, typically utilizing a multi-branch model for simultaneous prediction and leveraging both supervisions for training \cite{zhang2018recursive, xu2018pad, meng2019signet}.
Subsequent studies \cite{guizilini2020semantically, jung2021fine, zhu2023visual} have focused on improving the performance of one single task by exploiting information from the other.
For instance, Guizilini \textit{et al.} \cite{guizilini2020semantically} propose to use pre-trained segmentation networks to guide depth representation learning.
Jung \textit{et al.} \cite{jung2021fine} present a novel training method that exploits semantics-guided local geometry to optimize intermediate depth representations.

More recently, there have been efforts to explore joint learning of both sub-tasks in a unified manner.
MGNet \cite{schon2021mgnet} proposes a multi-task framework for monocular geometric scene understanding, which produces dense 3D point clouds with instance-aware semantic labels.
ViP-DeepLab \cite{qiao2021vip} extends Panoptic-DeepLab \cite{cheng2020panopticdeeplab} with an additional depth prediction head and introduces two datasets, along with an evaluation metric for the new task.
PanopticDepth \cite{gao2022panopticdepth} and PolyphonicFormer \cite{yuan2021polyphonicformer} propose unified architectures leveraging dynamic instance-specific kernels and query-based learning, respectively, to produce instance-level predictions.
Our approach has a unique advantage compared to previous works as it implicitly guides the learning of intermediate features through cross-modality supervision, and utilizes their mutual relations.

\section{Methods}

We present a unified framework for depth-aware panoptic segmentation where the panoptic segmentation and depth prediction are deeply unified through the learned latent representation.  
To further improve the joint learning of these two sub-tasks, we propose a bi-directional guided contrast learning approach that leverages cross-modality knowledge for more effective feature learning.

\begin{figure*}[t!]
    \centering
    \includegraphics[width=\textwidth]{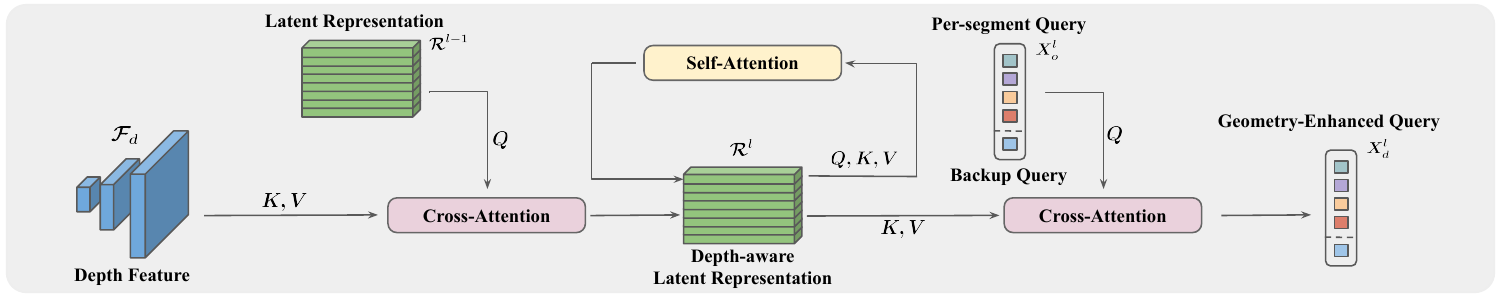}
    \caption{\textbf{Geometry Query Enhancement} with learned \textbf{Latent Representations}. We enhance per-segment queries with geometry information by operating masked cross-attentions and self-attentions alternatively, and set an extra \textbf{backup query} to cover up filtered-out regions in post-processing for better depth maps.}
    \label{depthquery}
\end{figure*}

\subsection{Network Architecture}

As shown in \cref{network}, our architecture is designed as an encoder-decoder structure with a shared encoder and two task-specific decoders.
The shared backbone extracts low-resolution features, and then two separate decoders gradually upsample features to generate semantic and depth feature pyramid, denoted as $\mathcal{F}$ and $\mathcal{F}_d$, respectively, with resolutions of $\times1/8$, $\times1/16$ and $\times1/32$.
Additionally, it includes $\times1/4$ pixel embedding $\mathcal{E}_{pixel}$ and depth embedding $\mathcal{E}_{depth}$.
For panoptic segmentation, we adopt the mask classification idea~\cite{cheng2021maskformer} due to its efficacy. Specifically, we utilize a Transformer Decoder with $l=9$ layers to process per-segment queries, where the input queries are sequentially interacted with the multi-scale semantic feature pyramid through 
masked attention~\cite{cheng2022mask2former}, giving rise to the output unified per-segment queries $X_o^l$. 
A multi-layer perceptron (MLP) followed by a Softmax layer is applied to the processed queries to generate the classification probability of all the segments. Meanwhile, the binary mask prediction is conducted through a dot product between the processed queries and pixel embedding, \ie, $M = \sigma ( \text{MLP} (X^l_o) \otimes \mathcal{E}_{pixel})$ with $\sigma$ indicating the Sigmoid activation.

\subsubsection{Per-Segment Depth Estimation}
\label{depthestimation}

We perform per-segment depth prediction in a similar way as the above panoptic segmentation process, allowing for a unified pipeline of both tasks. To further bridge the gap between the two tasks, we propose a geometric query enhancement module to incorporate geometry information into the unified per-segment queries.

\textbf{Geometric Query Enhancement.}
Instead of learning separate queries for different sub-tasks and linking them together as done in \cite{yuan2021polyphonicformer}, we manage to use unified queries and enhance them with geometry information, towards a deeply unified architecture.
Inspired by \cite{tri_define} and \cite{jaegle2021perceiverio}, we introduce a fixed-size latent representation (initialized as $\mathcal{R}^0$) to capture the global scene geometry, which serves as a middleware to communicate with per-segment queries and multi-scale depth features, as shown in \cref{depthquery}.
We first perform cross-attention between the masked depth features and the latent representation to project the geometry knowledge into the compressed latent space. 
To focus only on the regions of interest, we incorporate mask predictions from the segmentation branch, which results in faster convergence as shown in \cite{cheng2022mask2former}. 
Next, we apply self-attention in the latent space to update the latent representation $\mathcal{R}^{l}$. 
Finally, cross-attention is performed between the original per-segment queries $X^l_o$ and the updated latent representation $\mathcal{R}^{l}$ to generate the corresponding geometry-enhanced queries $X^l_{d}$.
In this way, each geometry-enhanced query is refined by the depth features to produce consistent per-segment depth maps.

\textbf{Depth Map Aggregation.}
Similar to mask predictions, we generate per-segment depth maps via dot product between the processed geometry-enhanced queries $X^l_d$ and depth embedding $\mathcal{E}_{depth} \in \mathbb{R}^{C_{d} \times \frac{H}{4} \times \frac{W}{4}}$, as
\begin{equation}
d = D_{max} \times \sigma ( \psi(X^l_d) \otimes \mathcal{E}_{depth} ),
\end{equation}
where $\sigma$ and $\psi$ denote the Sigmoid activation and feed-forward network (FFN) respectively, and $D_{max}$ is the max distance which is set to 80 in all experiments. The final depth map can be obtained by aggregating the per-segment depth according to the segmentation masks
\begin{equation}
    D(u,v) = \sum_{i\in \mathcal{H}} d_i(u,v) \cdot \mathbb{1}[M_i(u,v)>0.5],
\end{equation}
where $M_i$ denotes $i$-th segmentation mask, $\mathbb{1}[\cdot]$ denotes the indicator function, $(u,v)$ represents the spatial coordinate, and $\mathcal{H}$ only contains query ids with high-confidence segmentation masks\footnote{We follow the same post-processing as \cite{cheng2022mask2former} to filter out mask predictions with low confidence.}.

\textbf{Backup Query.}
In order to reduce the false positive rates in panoptic segmentation, low-confidence mask predictions are filtered out \cite{cheng2021maskformer}, but this may lead to blank segments in depth estimation.
Therefore, we introduce a backup query that produces a global depth map to address this issue.
Similar to the latent representations, the backup query uses cross-attention to update itself by querying multi-scale depth features, but without mask constraints, thereby enabling it to perceive global geometry knowledge instead of being limited to a specific region.
The resulting depth values, computed from the dot product of the backup query and the depth embedding $\mathcal{E}_{depth}$, can be utilized to supplement blank regions.
This simple idea not only enhances depth estimation performance but also mitigates the impact of inaccurate segmentation outcomes.

\subsection{Bi-directional Guidance Learning}
\label{bi}

Based on the tight-coupled relationship between semantic and geometry information, adjacent pixels that cross semantic boundaries are more likely to have large differences in depth, and vice versa.
The cross-domain knowledge learning has been employed in several methods \cite{guizilini2020semantically, jung2021fine}, mainly by leveraging scene semantics to produce semantically consistent intermediate depth representations.
However, these one-way methods only guide the learning of depth representations from semantic labels but not the other way around.
Therefore, we propose a bi-directional guidance learning method to refine both semantic and depth representations simultaneously, which not only boosts performance on both sub-tasks but also achieves improvement under incomplete supervision, where the ground truth for only one task is available.
The learning process occurs only during training, and thus no additional computation is required during inference.

\subsubsection{Semantic-to-Depth Guidance}
\begin{figure}[t]
\centering
\includegraphics[width=\columnwidth]{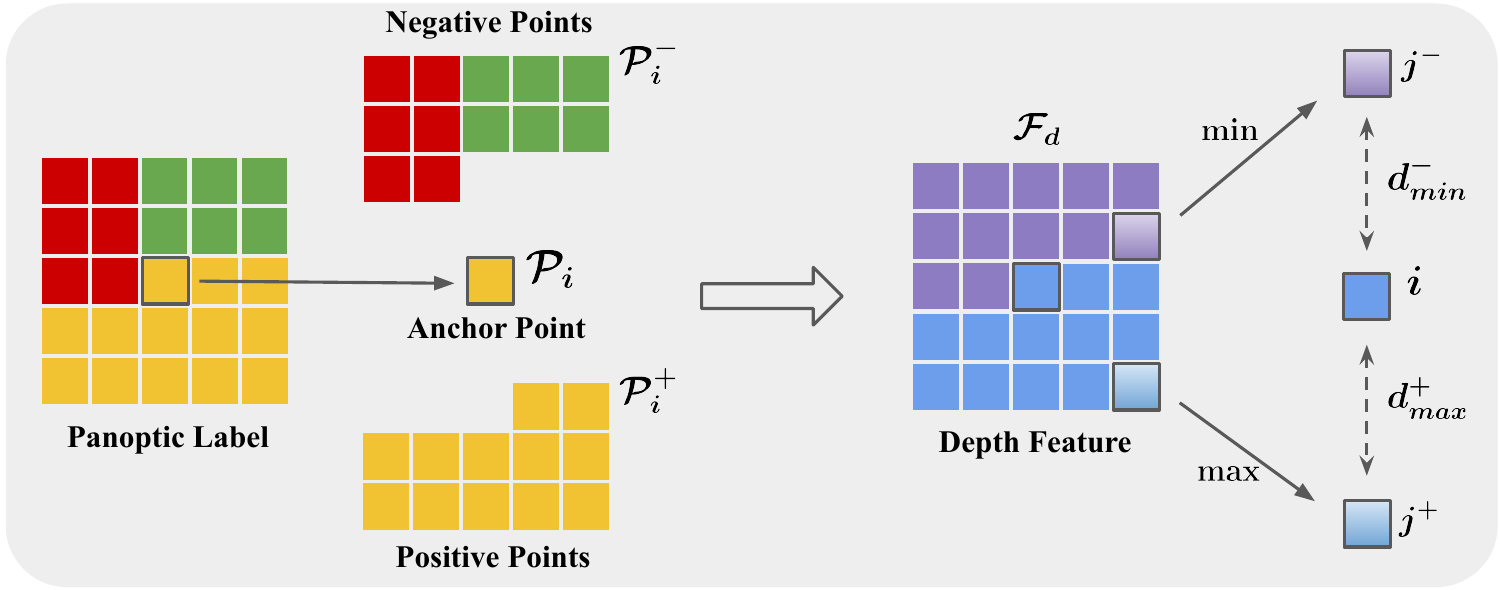}
\caption{
Overview of Semantic-to-Depth Guidance. 
We optimize relative depth feature distances following a max-min strategy, inside each $K \times K$ local patch.
}
\label{sem}
\end{figure}

We use contrastive learning to refine depth representations, which aims to minimize the feature distance between pixels within the same instance while maximizing those across instance boundaries.
This ensures that depth features are more discriminating on boundary regions, consistent with semantic information.
Inspired by prior works \cite{jung2021fine, guizilini2020semantically}, we constrain the learning process to local patches since pixels far away from each other may not meet the geometry consistency assumption.
In contrast to \cite{jung2021fine}, we employ a max-min strategy to enforce the maximum feature distance within the same instance to be smaller than the minimum distance of different instance, with only two pixels being selected for contrastive learning (illustrated in \cref{sem}).

First, we divide the panoptic labels into patches of size $K \times K$ with a stride of $1$, and select the center point of each patch as the anchor point $\mathcal{P}_i$.
Subsequently, we consider points with the same panoptic label as the anchor point to be positive points $\mathcal{P}_i^+$, and all other points as negative points $\mathcal{P}_i^-$.
To exclude patches that do not contain instance boundaries, we only select patches with $|\mathcal{P}_i^-| > 0$.
We define the maximum positive distance $d^+_{max}$ and minimum negative distance $d^-_{min}$ as the L2 distance of the normalized depth feature, which are formulated as
\begin{align}
    d^+_{max}(i) &= \max (\| \hat{\mathcal{F}}^l_d(i) - \hat{\mathcal{F}}^l_d(j^+) \|_2), \, j^+ \in \mathcal{P}^+_i \\
    d^-_{min}(i) &= \min (\| \hat{\mathcal{F}}^l_d(i) - \hat{\mathcal{F}}^l_d(j^-) \|_2), \, j^- \in \mathcal{P}^-_i,
\end{align}
where $l$ denotes the index of feature layer and $\hat{\mathcal{F}_d}$ denotes the normalized depth features, \textit{i.e.}, $\hat{\mathcal{F}_d} = \mathcal{F}_d / \| \mathcal{F}_d \|$.

We aim to enforce the learning process towards the objective of $d^+_{max}(i) < d^-_{min}(i)$, following the intuition that the maximum depth difference within an instance is likely to be smaller than the minimum depth difference across the boundary of the instance region.
Our semantic guidance loss embodies the above intuition and adopts the following triplet-based form~\cite{wang2014ranking}:
\begin{equation}
    \mathcal{L}_{sg} = \frac{1}{N} \sum_i \max(0, \alpha + d^+_{max}(i) - d^-_{min}(i)),
\end{equation}
where $\alpha$ is a gap parameter that regularizes the margin between the two distances, $i$ only includes patches that contain boundaries, and $N$ is the number of such patches.
The final semantic guidance loss is averaged across multiple layers.

\subsubsection{Depth-to-Semantic Guidance}

Following a similar idea, we in turn use depth supervision to guide the learning of semantic representations.
Unlike panoptic labels, depth annotations cannot be simply grouped into different segments.
Hence, we aim to enforce the continuity consistency between the depth supervision and intermediate semantic feature representations, based on the observation that pixels with continuously varying depth values are usually located within the same instance, while dramatic discontinuity usually occurs around instance boundaries.
Considering the local geometry consistency, we restrict the learning process to local patches as well and choose the same patch size $K$ as the semantic guidance learning.

Inside each patch, we choose the center pixel $i$ as the reference pixel and all the rest as neighboring pixels. Then we calculate, for each neighboring pixel $j$, its relative depth distance and semantic feature distance from the reference pixel.
The depth guidance loss is defined as
\begin{equation}
    \mathcal{L}_{dg}(i,j) = -\frac{1}{N} \sum_i \sum_j e^{- \| \hat{d}_i - \hat{d}_j \| /\tau} \cdot e^{-\| \mathcal{F}_i - \mathcal{F}_j \|_2}
\end{equation}
where $\hat{d}$ and $\mathcal{F}$ denote the groundtruth depth and semantic features, respectively, and $\tau$ is the scaling factor to balance the difference in magnitude between two distances, which is set to 10 throughout the experiment.
$i$ and $j$ only contain pixels whose depth annotations are available, and $N$ is the number of available patches.
The final depth guidance loss is also averaged across multiple layers.

\begin{table*}[t]
\resizebox{\textwidth}{!}{%
\begin{tabular}{l|cc|ccc|ccc|ccc|ccc|c}
\toprule
\bf{Cityscapes-DVPS} & Backbone & Extra Data & \multicolumn{3}{|c|}{$\lambda=0.5$} & \multicolumn{3}{|c|}{$\lambda=0.25$} & \multicolumn{3}{|c|}{$\lambda=0.1$} & \multicolumn{3}{|c|}{DPQ} & FLOPs \\ 
\midrule
PanopticDepth\cite{gao2022panopticdepth} & ResNet-50 & - & 65.6 & 59.2 & 70.2 & 62.3 & 57.0 & 66.1 & 43.2 & 40.7 & 45.1 & 57.0 & 52.3 & 60.5 & 619G \\
PolyphonicFormer\cite{yuan2021polyphonicformer} & ResNet-50\textasteriskcentered & MV & 64.3 & 56.0 & 70.3 & 59.7 & 53.3 & 64.4 & 39.3 & 31.8 & 44.7 & 54.4 & 47.0 & 59.8 & - \\
\bf{Ours} & ResNet-50 & - & \bf{69.3} & \bf{61.4} & \bf{75.0} & \bf{66.8} & \bf{59.1} & \bf{72.4} & \bf{52.8} & \bf{46.9} & \bf{57.1} & \bf{63.0} & \bf{55.8} & \bf{68.2} & 510G \\

\hline
ViP-DeepLab$\dagger$\cite{qiao2021vip} & WR-41\textasteriskcentered & MV, CSV & 68.7 & 61.4 & 74.0 & 66.5 & 60.4 & 71.0 & {50.5} & {45.8} & 53.9 & 61.9 & {55.9} & 66.3 & 4,725G \\
PolyphonicFormer\cite{yuan2021polyphonicformer} & Swin-B\textasteriskcentered & MV & \bf{70.6} & \bf{63.0} & \bf{76.0} & {67.8} & {61.0} & {72.8} & 50.2 & 43.4 & {55.2} & {62.9} & 55.8 & {68.0} & 837G \\
\bf{Ours} & Swin-B & - & 69.8 & 62.3 & 75.3 & \bf{68.1} & \bf{61.4} & \bf{73.0} & \bf{55.0} & \bf{48.7} & \bf{59.5} & \bf{64.3} & \bf{57.5} & \bf{69.3} & 1,037G \\
\bottomrule
\end{tabular}%
}

\resizebox{\textwidth}{!}{%
\begin{tabular}{l|cc|ccc|ccc|ccc|ccc|c}
\toprule
\bf{SemKITTI-DVPS} & Backbone & Extra Data & \multicolumn{3}{|c|}{$\lambda=0.5$} & \multicolumn{3}{|c|}{$\lambda=0.25$} & \multicolumn{3}{|c|}{$\lambda=0.1$} & \multicolumn{3}{|c|}{DPQ} & FLOPs \\ 
\midrule
PanopticDepth\cite{gao2022panopticdepth} & ResNet-50 & - & - & - & - & - & - & - & - & - & - & 46.9 & \bf{46.0} & 47.6 & 144G \\
PolyphonicFormer\cite{yuan2021polyphonicformer} & ResNet-50\textasteriskcentered & MV & 50.5 & 44.0 & 55.3 & 47.9 & 42.2 & 52.1 & 35.9 & 33.6 & 37.6 & 44.8 & 39.9 & 48.3 & - \\
\bf{Ours} & ResNet-50 & - & \bf{54.7} & \bf{48.8} & \bf{59.0} & \bf{51.4} & \bf{46.5} & \bf{54.9} & \bf{37.7} & \bf{34.0} & \bf{40.4} & \bf{47.9} & 43.1 & \bf{51.5} & 164G \\
\hline
ViP-DeepLab$\dagger$\cite{qiao2021vip} & WR-41\textasteriskcentered & MV, CSV & 54.7 & 46.4 & 60.6 & 52.0 & 44.8 & 57.3 & 40.0 & 34.7 & \bf{43.8} & 48.9 & 42.0 & \bf{53.9} & 1,133G \\
PolyphonicFormer\cite{yuan2021polyphonicformer} & Swin-B\textasteriskcentered & MV & {58.5} & {55.1} & {61.0} & \bf{56.3} & {54.0} & \bf{57.9} & {41.8} & {41.1} & 42.4 & {52.2} & {50.1} & 53.8 & 201G \\
\bf{Ours} & Swin-B & - & \bf{59.7} & \bf{57.1} & \bf{61.5} & \bf{56.3} & \bf{54.8} & {57.4} & \bf{42.4} & \bf{42.0} & {42.8} & \bf{52.8} & \bf{51.3} & \bf{53.9} & 281G \\
\bottomrule
\end{tabular}%
}
\caption{Depth-aware panoptic segmentation results on Cityscapes-DVPS and SemKITTI-DVPS. 
`MV': Mapillary Vistas \cite{neuhold2017mv}. 
`CSV': Cityscapes videos with pseudo labels \cite{chen2020naive}. 
\textdagger: test-time augmentation.
\textasteriskcentered: Recursive Feature Pyramid (RFP) \cite{qiao2021rfp}.
Each cell shows $\text{DPQ}^\lambda$ \textbar $\text{DPQ}^\lambda$-Thing \textbar $\text{DPQ}^\lambda$-Stuff, where $\lambda$ is the threshold of relative depth error. 
}
\label{table:citydvps}
\end{table*}

\subsection{Losses}

Following prior works \cite{cheng2021maskformer, cheng2022mask2former}, we first find one-to-one matching between per-segment queries and ground-truth masks via bipartite matching, then adopt the same loss as \cite{cheng2022mask2former}: a cross-entropy classification loss $\mathcal{L}_{cls}$, and a combination of binary cross-entropy loss and dice loss \cite{milletari2016dice} for mask predictions: $\mathcal{L}_{mask} = \mathcal{L}_{ce} + \mathcal{L}_{dice}$.

As our geometry-enhanced queries correspond to per-segment queries as well, the depth predictions are assigned with the same matching results.
We use the scale-invariant loss \cite{eigen14}, which is simple yet efficient, and formulated as
\begin{equation}
    \mathcal{L}_{depth} = \frac{1}{n} \sum_i g_i^2 - \frac{\lambda}{n^2} (\sum_i g_i )^2
\end{equation}
where $g_i = \log d_i - \log \hat{d_i}$ with the predicted depth $d$ and ground-truth depth $\hat{d}$, and $\lambda$ is set to $0.85$.

The total loss is the weighted sum of all loss terms, as
\begin{equation}
\begin{split}
    \mathcal{L} = \lambda_{cls} \mathcal{L}_{cls} & + \lambda_{mask} \mathcal{L}_{mask} + \lambda_{depth} \mathcal{L}_{depth} \\
    &+ \lambda_{sg} \mathcal{L}_{sg} + \lambda_{dg} \mathcal{L}_{dg},
\end{split}
\end{equation}
where we use $\lambda_{cls}=2$, $\lambda_{mask}=5$, $\lambda_{depth}=2.5$ and $\lambda_{sg}=\lambda_{dg}=0.1$ in our experiments.

\section{Experiments}

\subsection{Datasets}

{\textbf{Cityscapes-DVPS}} \cite{qiao2021vip} is an extension of Cityscapes-VPS \cite{kim2020vps} that includes depth annotations converted from disparity maps using stereo images.
It consists of 3,000 annotated frames, with 2,400, 300, and 300 frames in the training, validation, and test sets, respectively. 
The dataset maintains the same semantic classes as the original Cityscapes \cite{cordts2016cityscapes} dataset, which include 8 thing and 11 stuff classes.

{\textbf{SemKITTI-DVPS}} \cite{qiao2021vip} is derived from the odometry split of the KITTI dataset \cite{geiger2012kitti}.
The dataset includes 11 training sequences, 11 test sequences, and validation is performed on sequence 08. 
Sparse semantic annotations are obtained by projecting panoptic-labelled 3D point clouds from SemanticKITTI \cite{behley2019semantickitti} onto the image plane. The dataset comprises 19,130 training images, 4,071 evaluation images, and 4,342 test images.

\subsection{Evaluation Metrics}

We evaluate results using standard evaluation metrics, with Panoptic Quality (PQ) \cite{kirillov2019panoptic} for panoptic segmentation and Depth-aware Panoptic Quality (DPQ) for depth-aware panoptic segmentation introduced by \cite{qiao2021vip}.
Specifically, let $P$ and $Q$ be the prediction and ground-truth, we use $P$ and $P^d$ to denote segmentation and depth estimation respectively, and the same notation also applies to $Q$.
Then, $DPQ^\lambda$ is defined as
$$ DPQ^\lambda (P, Q) = PQ(P^\lambda, Q) ,$$
where $P^\lambda = P$ for pixels that have absolute relative depth errors under the threshold $\lambda$ (\textit{i.e.}, $|P^d - Q^d| \leq \lambda Q^d$), and other pixels will be assigned to \textit{void}. $DPQ^\lambda$ is evaluated under three values of $\lambda = \{0.1,0.25,0.5\}$, and averaged to obtain the final DPQ.

\begin{figure*}[t]
\centering
\includegraphics[width=\textwidth]{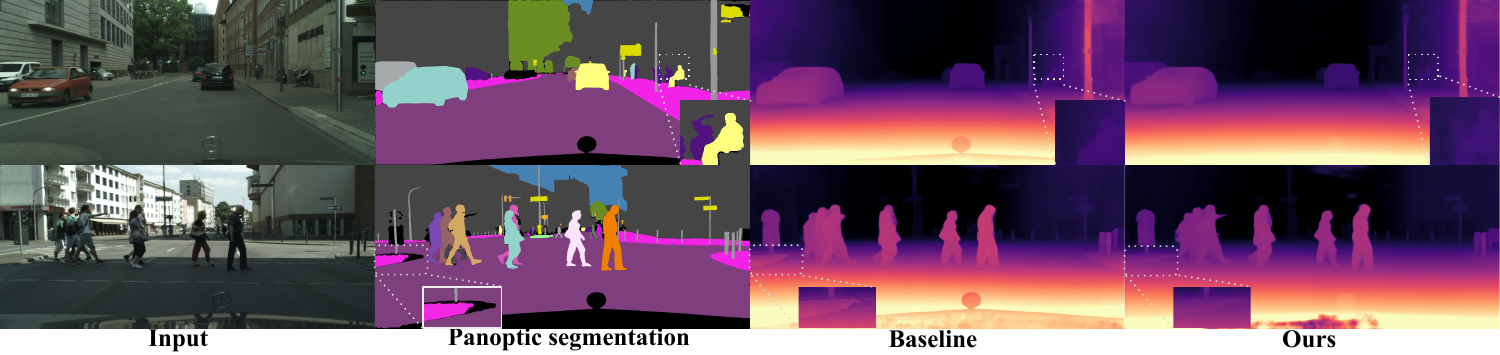}
\caption{Visualization results on Cityscapes-DVPS. 
Top row: unified architecture distinguishes boundaries for better depth estimation. 
Bottom row: backup query alleviates the case of imperfect mask predictions.
}
\label{fig:visual}
\end{figure*}%

\subsection{Implementation}

We use Detectron2 \cite{wu2019detectron2} to implement our model and follow \cite{cheng2022mask2former} to choose multi-scale deformable attention Transformer \cite{zhu2020deformable} as multi-scaled decoders.
We adopt ResNet-50 \cite{he2016resnet} and Swin-B \cite{liu2021swin} as the shared backbone, and do not use extra datasets for pre-training or Recursive Feature Pyramid (RFP) \cite{qiao2021rfp} to enhance the backbone, which differs from prior methods \cite{qiao2021vip, yuan2021polyphonicformer}.

Following prior works \cite{gao2022panopticdepth, yuan2021polyphonicformer}, the full training process consists of two steps. 
First, we train the segmentation branch using panoptic labels only for 50 epochs. 
Then, we finetune the entire model with both supervisions for 10 epochs.
During the first step, we resize images with a random scale from 0.5 to 2.0, followed by a fixed size crop to $512 \times 1024$ and $384 \times 1280$ on Cityscapes-DVPS and SemKITTI-DVPS, respectively.
After the pre-training phase, we use full-resolution for finetuning.
Large-scale jittering (LSJ) \cite{liu2016ssd} and horizontal flipping are also employed during training.
Final predictions are obtained from a single inference, while no test-time augmentation is employed.
More details are provided in supplementary material.

\subsection{Main Results}

\begin{table}[t]
\small
\centering
\resizebox{\columnwidth}{!}
{%
\begin{tabular}{l|c|c|ccc}
\toprule
Method & Backbone & Depth & PQ & PQ$^{\text{Th}}$ & PQ$^{\text{St}}$ \\ 
\midrule
VPSNet \cite{kim2020vps} & ResNet-50 &  & 65.0 & - & - \\
ViP-DeepLab \cite{qiao2021vip} & ResNet-50 & $\checkmark$ & 60.6 & - & -  \\
PanopticDepth\textasteriskcentered \cite{gao2022panopticdepth} & ResNet-50 & $\checkmark$ & 66.9 & 60.1 & 71.9\\
PolyphonicFormer \cite{yuan2021polyphonicformer} & ResNet-50 & $\checkmark$ & 65.4 & - & -  \\
\bf{Ours} & ResNet-50 & $\checkmark$ & \bf{69.5} & \bf{62.1} & \bf{75.7} \\
\bottomrule
\end{tabular}
}%
\caption{
Panoptic Segmentation results on Cityscapes-DVPS validation set. 
}
\label{table:pq}
\end{table}

\begin{table}[t]
\centering
\resizebox{\columnwidth}{!}
{%
\begin{tabular}{l|cc|ccc}
\toprule
Method & abs rel & RMSE log & $\sigma < 1.25$ & $\sigma < 1.25^2$ & $\sigma < 1.25^3$  \\ 
\midrule
DPT-Hybrid \cite{ranftl21dpt} & 0.0697 & 0.1106 & 0.9434 & 0.9914 & 0.9976 \\
PanopticDepth \cite{gao2022panopticdepth} & 0.0711 & 0.1125 & 0.9359 & 0.9919 & 0.9982  \\
PolyphonicFormer \cite{yuan2021polyphonicformer} & 0.0647 & {0.1013} & 0.9524 & 0.9950 & {0.9985}  \\
\bf{Ours} & \bf{0.0597} & \bf{0.0940} & \bf{0.9616} & \bf{0.9953} & \bf{0.9988} \\
\bottomrule
\end{tabular}
}%
\caption{Monocular Depth Estimation results on Cityscapes-DVPS validation set.}
\label{table:depth}
\end{table}

\textbf{Depth-aware Panoptic Segmentation.}
We evaluate our model on two datasets, Cityscapes-DVPS and SemKITTI-DVPS, and compare it with recent methods.
Our method achieves state-of-the-art results on both datasets as presented in \cref{table:citydvps}.
It is worth mentioning that we did not employ any additional techniques, such as pre-training on larger datasets \cite{neuhold2017mv}, training with pseudo-labels \cite{chen2020naive}, backbone enhancement \cite{qiao2021rfp}, or test-time augmentation \cite{qiao2021vip}.

On Cityscapes-DVPS, our model achieves superior performance compared to published methods that use larger backbones (WR-41 \cite{Zagoruyko2016WRN} and Swin-B \cite{liu2021swin} with RFP \cite{qiao2021rfp}) despite using the less powerful ResNet-50 \cite{he2016resnet} backbone.
Although our model performs slightly worse than PolyphonicFormer \cite{yuan2021polyphonicformer} with Swin-B on $\text{DPQ}^{0.5}$, it demonstrates significant improvement on $\text{DPQ}^{0.1}$, indicating the better depth quality, especially on challenging thresholds.

Learning on SemKITTI-DVPS is more challenging, because both ground-truths are sparse, leading to much fewer available training pixels. 
\cref{table:citydvps} reports our DPQ result on SemKITTI-DVPS validation set, and we outperform all state-of-the-art methods under the same backbone.

\textbf{Individual Sub-tasks.}
We have evaluated our model on two individual sub-tasks and presented the results in tables \cref{table:pq} and \cref{table:depth}.
Our approach outperforms state-of-the-art methods on both sub-tasks, indicating that the unified architecture and cross-modality learning process are advantageous for both sub-tasks.

\textbf{Visualization Results.}
The panoptic segmentation and depth estimation results on Cityscapes-DVPS are visualized in \cref{fig:visual}.
The top row shows the benefit of the unified framework, which improves depth results with clearer boundaries, especially on small objects (\textit{e.g.}, bicycle in the distance).
The bottom row shows that the backup query could provide smoother depth values in blank mask regions, caused by imperfect mask predictions.

\begin{figure}[t]
\centering
\includegraphics[width=\columnwidth]{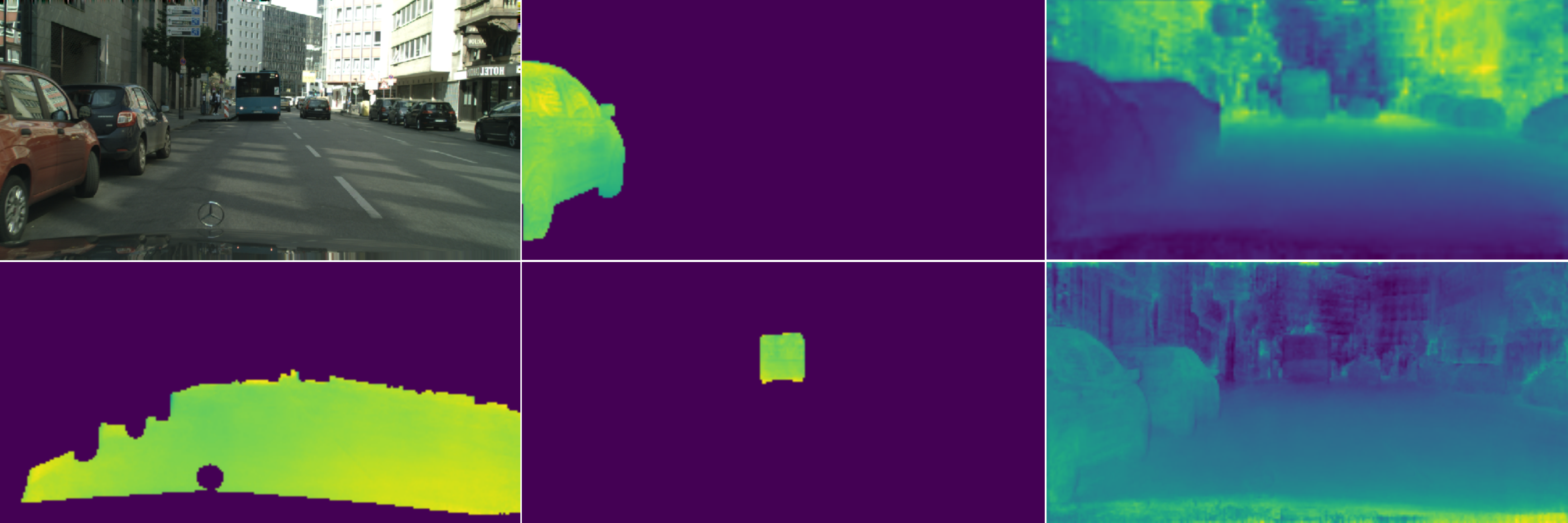}
\caption{Latent representation attention maps.
}
\label{fig:attn}
\end{figure}

In addition, we choose 5 latent representation indices and visualize their attention maps in \cref{fig:attn}.
We discover that certain latent representations specialize in various instances such as roads (bottom left), cars and buses (middle column), as well as different distance ranges (right cloumn).
More visualization results are available in the supplementary materials.

\subsection{Ablation Studies}

\begin{table*}[t]
\resizebox{\textwidth}{!}
{%
\begin{tabular}{c|cccc|cccc|cc|c}
\toprule
Variants & Instance-wise & Backup Query & S.Guidance & D.Guidance & $\lambda=0.5$ & $\lambda=0.25$ & $\lambda=0.1$ & DPQ & PQ $\uparrow$ & abs rel $\downarrow$ & \#comp. \\ 
\midrule
A &  &  & & & 68.5 & 64.3 & 48.3 & 60.3 & 69.0 & 0.087 & - 8.6\% \\
B & \checkmark &  &  & & 68.7 & 66.0 & 51.8 & 62.1 & 68.9 & 0.068 & -  \\
C & \checkmark & \checkmark & &  & 68.9 & 66.5 & 51.9 & 62.4 & 69.0 & 0.065 & + 4.2\% \\
D & \checkmark & \checkmark & \checkmark & & \bf{69.3} & \bf{67.1} & 51.9 & 62.8 & \bf{69.5} & 0.064 & + 17.8\% \\ 
E & \checkmark & \checkmark & \checkmark & \checkmark & \bf{69.3} & {66.8} & \bf{52.8} & \bf{63.0} & \bf{69.5} & \bf{0.063} & + 19.1\% \\
\bottomrule
\end{tabular}
}%
\caption{Ablation studies on Cityscapes-DVPS. 
`Instance-wise': unified architecture instead of attaching a depth regression head to the baseline model. 
`S/D.Guidance': Semantic-to-Depth/Depth-to-Semantic Guidance Learning.}
\label{table:ablation}
\end{table*}

\textbf{Depth-aware Panoptic Segmentation.}
To demonstrate the advantages of our proposed architecture and learning method, we conducted ablation studies on the Cityscapes-DVPS dataset in \cref{table:ablation}, with ResNet-50 backbone and full resolution images. 
Our baseline model, Variant-A, simply adds an additional CNN depth regression head to the segmentation branch.
By replacing the basic depth head with the proposed instance-wise depth decoder, our model achieves a 1.8\% improvement, and Backup Query further enhances it by 0.3\%.
Interestingly, we observed that the guidance learning process not only facilitates cross-domain feature learning but also benefits itself. 
We speculate that the self-improvement stems from the joint-learning framework, where both queries communicate with each other through intermediate latent representations.
We also provide analysis of computation costs associated with different components (\#comp. in \cref{table:ablation}), and found that less than 20\% extra training cost brings boosting performance.

Another notable advantage we have observed over the previous method \cite{qiao2021vip} is that, hyper-parameters have a limited effect on the final performance, which is consistent with the findings in \cite{yuan2021polyphonicformer}, thanks to the unified framework.
Therefore, we keep most hyper-parameters the same as in previous works (\textit{e.g.}, segmentation loss weights \cite{cheng2022mask2former}, gap parameter $\alpha$ and patch size $K$ \cite{jung2021fine}).
More hyper-parameter ablations can be found in supplementary.

\begin{table}[t]
\centering
\resizebox{\columnwidth}{!}
{%
\begin{tabular}{l|cccc|cc}
\toprule[1pt]
 & $\lambda=0.5$ & $\lambda=0.25$ & $\lambda=0.1$ & DPQ & PQ $\uparrow$ & abs rel $\downarrow$ \\ 
\midrule
Full-supervision & 67.6 & 65.2 & 42.7 & 58.5 & 67.8 & 0.112\\
\hline
Semi-supervision & 66.1 & 63.3 & 37.9 & 55.8 & 66.4 & 0.140 \\
+ depth guidance & 67.0 & 64.5 & 39.2 & 56.9 & 67.1 & 0.131 \\
+ semantic guidance & 66.6 & 64.7 & \bf{41.7} & 57.7 & 66.7 & \bf{0.122} \\
+ \bf{both guidance} & \bf{67.1} & \bf{65.0} & 41.2 & \bf{57.8} & \bf{67.4} & 0.126 \\
\bottomrule[1pt]
\end{tabular}
}%
\caption{Effect of Bi-directional Guidance Learning. 
`Semi-supervision': incomplete ground-truth labels.
}
\label{table:bi}
\end{table}

\begin{table}[t]
\centering
\resizebox{\columnwidth}{!}
{%
\begin{tabular}{l|cccc|cc}
\toprule[1pt]
 & $\lambda=0.5$ & $\lambda=0.25$ & $\lambda=0.1$ & DPQ & PQ $\uparrow$ & abs rel $\downarrow$ \\ 
\midrule
w/o guidance & 67.5 & 63.9 & 41.2 & 57.5 & 67.7 & 0.132 \\
\hline
SGT \cite{jung2021fine} & \bf{67.5} & 64.0 & 42.5 & 58.0 & 68.0 & 0.120 \\
\bf{ours S.Guide} & \bf{67.5} & \bf{64.3} & \bf{42.9} & \bf{58.2} & \bf{68.1} & \bf{0.117} \\
\hline
DGS & 67.6 & 64.2 & {42.3} & 58.0 & 68.3 & {0.129} \\
\bf{ours D.Guide} & \bf{67.7} & \bf{64.3} & \bf{42.6} & \bf{58.2} & \bf{68.5} & \bf{0.126} \\
\bottomrule[1pt]
\end{tabular}
}%
\caption{
Ablation studies on choices of Bi-directional Guidance Learning losses.
}
\label{table:loss}
\end{table}

\begin{table}[t]
\resizebox{\columnwidth}{!}
{%
\begin{tabular}{l|cccc|cc}
\toprule[1pt]
 & $\lambda=0.5$ & $\lambda=0.25$ & $\lambda=0.1$ & DPQ & PQ $\uparrow$ & abs rel $\downarrow$ \\
\midrule
Single query & 65.8 & 59.3 & 34.8 & 53.3 & 66.4 & 0.163 \\
Double latent & 67.4 & 64.3 & 40.4 & 57.4 & {67.8} & 0.119 \\
Query Linking \cite{yuan2021polyphonicformer} & 67.5 & 63.9 & 41.2 & 57.5 & 67.8 & 0.123 \\
\hline
\bf{Latent representation} & \bf{67.6} & \bf{65.2} & \bf{42.7} & \bf{58.5} & \bf{67.8} & \bf{0.112} \\
\bottomrule
\end{tabular}
}%
\caption{Effect of Geometric Query Enhancement.}
\label{table:depthquery}
\end{table}

\textbf{Bi-directional Guidance Learning.}
To further demonstrate the effectiveness of the proposed bi-directional learning method, we conduct extra experiments by simulating training with incomplete supervision labels.
We separate the full training data into three subsets, where the first subset only contains panoptic labels, the second subset only contains depth annotations, and the last one keeps both annotations unchanged.
All experiments are trained on $512 \times 1024$ images using ResNet-50 backbone.
As shown in \cref{table:bi}, we find that training with incomplete supervision leads to dramatic performance drops (-3.3\%) on DPQ and both sub-tasks as well.
Introducing the proposed guidance loss terms during training results in performance boosts and achieves minimum performance gap with full supervision using bi-directional guidance.

We additionally show comparisons of different forms of guidance loss design in \cref{table:loss}.
``SGT" refers to semantic-guided triplet loss proposed in \cite{jung2021fine}, and ``DGS" refers to disparity-guided smoothness, which regularizes the smoothness of semantic features based on disparity gradients (vertically and horizontally), modified from \cite{chen2019towards}.
We find that any implementation of such guidance learning is beneficial, and experimentally choose the best one.
The result also indicates that the idea of bi-directional guidance learning is more important than the loss design choice.

\textbf{Geometric Query Enhancement.}
We conduct several experiments to generate queries for depth estimation and show results in \cref{table:depthquery}.
``Single query" denotes using the same per-segment queries to incorporate both features. 
``Double latent" denotes adding an extra latent representation for semantic features.
``Query Linking" denotes the method in \cite{yuan2021polyphonicformer}, which adopts grouping to select salient features followed by query update and reasoning.
We found the geometric-enhanced queries with learned latent representations achieve the best performance among all alternatives.

\subsection{Limitations}

As shown in \cref{fig:visual}, inaccurate mask predictions affect the quality of the depth map, making decent pre-training for segmentation necessary for follow-up depth learning. 
This suggests that the model may not generalize well to other scenes with unseen objects.
In the future, we hope to address the above limitations and improve the model towards the goal of a fully unified framework.

\section{Conclusion}
We present a deeply unified framework for depth-aware panoptic segmentation at both the architectural and learning levels, exploiting the natural correlation between these two sub-tasks.
By leveraging cross-modality guidance learning, the intermediate feature representations not only benefit from each other but also from themselves in turn.
With the above contributions, the proposed method achieves state-of-the-art results on two datasets and shows promise in improving performance even with incomplete supervision labels.
We hope our approach will inspire future researches in geometric scene understanding.

\noindent \textbf{Acknowledgements.}
The paper is supported in part by the National Key R\&D Program of China (2018AAA0102001), National Natural Science Foundation of China (62276045, 62006036, 62293542, U1903215), Dalian Science and Technology Talent Innovation Support Plan (2022RY17), and Fundamental Research Funds for Central Universities (DUT22LAB124, DUT22ZD210, DUT22QN228, DUT21RC(3)025).

{\small
\bibliographystyle{ieee_fullname}
\bibliography{egbib}
}

\end{document}